\documentclass[conference]{IEEEtran}
\IEEEoverridecommandlockouts

\usepackage{cite}
\usepackage{amsmath,amssymb,amsfonts}
\usepackage{algorithmic}
\usepackage{graphicx}
\usepackage{textcomp}
\usepackage{float}
\usepackage{xcolor}
\def\BibTeX{{\rm B\kern-.05em{\sc i\kern-.025em b}\kern-.08em
    T\kern-.1667em\lower.7ex\hbox{E}\kern-.125emX}}
\begin{document}

\title{Cost-Effective Robotic Handwriting System with AI Integration}

\author{\IEEEauthorblockN{Tianyi Huang}
\IEEEauthorblockA{\textit{App-In Club} \\
Fremont, California, USA \\
tonyhrule666@gmail.com}
\and
\IEEEauthorblockN{Richard Xiong}
\IEEEauthorblockA{\textit{App-In Club} \\
Trumbull, Connecticut, USA \\
richardx366@gmail.com}
}

\IEEEpubid{\begin{minipage}{\textwidth}\ \\[12pt]
  978-8-3315-0666-7/\$31.00 \copyright 2024 IEEE\\ 
  2024 IEEE Long Island Systems, Applications and Technology Conference
\end{minipage}} 

\maketitle

\begin{abstract}
This paper introduces a cost-effective robotic handwriting system designed to replicate human-like handwriting with high precision. Combining a Raspberry Pi Pico microcontroller, 3D-printed components, and a machine learning-based handwriting generation model implemented via TensorFlow.js, the system converts user-supplied text into realistic stroke trajectories. By leveraging lightweight 3D-printed materials and efficient mechanical designs, the system achieves a total hardware cost of approximately \$56, significantly undercutting commercial alternatives. Experimental evaluations demonstrate handwriting precision within $\pm$0.3 millimeters and a writing speed of approximately 200 mm/min, positioning the system as a viable solution for educational, research, and assistive applications. This study seeks to lower the barriers to personalized handwriting technologies, making them accessible to a broader audience.

\end{abstract}


\section{Introduction}

Despite the increasing popularity of digital communications, handwritten text retains a unique and sentimental value in both personal and professional contexts \cite{tanner2019education,calvet2021lowcost}. Traditional handwriting machines, however, are often expensive, complex, and bulky, limiting their accessibility primarily to large companies \cite{gu2019lowcost,cheng2019rapid}. These systems typically cost around \$150 and use expensive materials, posing challenges in terms of cost and usability for smaller groups or individual users \cite{sas2021affordable}.

To address these challenges, this study introduces a cost-effective robotic handwriting system that uses a Raspberry Pi Pico microcontroller and 3D-printed components to reduce the overall cost to approximately \$56 per unit \cite{patel2021pico}. This cost reduction allows for enhanced accessibility for educational settings and small businesses where such technology was previously considered too expensive \cite{kim2019cost}.

The Raspberry Pi Pico provides a compact yet powerful platform for controlling the mechanical components of the handwriting machine. Combined with 3D printing technology for structural components, the system allows for easy prototyping and customization, enhancing its adaptability \cite{cheng2019rapid,calvet2021lowcost}. By replacing traditional metal components with lightweight 3D-printed plastics and simpler mechanical designs such as lead screws instead of timing belts, the system becomes cheaper to produce and more energy-efficient \cite{intarapanich2020comparison}.

In addition to its mechanical innovations, this paper incorporates a machine learning-based handwriting generation model implemented via TensorFlow.js \cite{smilkov2019tfjs}. This AI integration enables the system to convert user-provided text into stroke trajectories that emulate human-like writing \cite{graves2013generating,ha2017neural}. By doing so, this system enhances the realism and personalization of the handwriting output, expanding its utility for creative, educational, and assistive applications \cite{chen2020personalized}. Experimental results confirm that the system achieves handwriting precision within $\pm$0.3 millimeters while maintaining efficient operations and low costs.

This paper explains the design and implementation of the overall system, discusses its components, provides cost analysis, and presents experimental results that demonstrate its accuracy and effectiveness. We also explore potential applications and future work to improve the system further.

\section{System Implementation}
\subsection{Overview of Development Process}
The development of the hardware system involved four major stages:
\begin{enumerate}
    \item \textbf{Mechanical Design:} Using Autodesk Fusion 360, a 3D CAD design software, the entire mechanical framework was carefully designed \cite{autodeskFusion3602019}. This included the structural frame, lead screw mechanisms for movement along the X, Y, and Z axes, the pen holder, and mounts for electronic components. Precise modeling ensured that all parts would function correctly upon assembly.
    \item \textbf{Assembly:} The components were made using a Creality Ender 3 3D printer with PLA filament \cite{creality2019ender3}. The 3D-printed parts were then assembled along with the specified hardware components, including the Raspberry Pi Pico microcontroller, 28BYJ-48 stepper motors, ULN2003 driver boards, limit switches, and ball bearings \cite{28BYJ48DataSheet,uln2003DriverDataSheet}. Assembly involved securing components through friction fit or screws.
    \item \textbf{Hardware Controls:} The Raspberry Pi Pico was programmed using an Arduino sketch with a modified version of the AccelStepper library \cite{arduinoIDE,accelStepperLib}. The front-end interface was developed using HTML, CSS, and JavaScript, incorporating TensorFlow.js for handwriting generation \cite{malaviya2020tensorFlowjs}. The microcontroller code was uploaded via USB for direct communication with the controlling computer.
    \item \textbf{Testing:} Movements to specific coordinates were tested to verify accuracy. Calibration involved adjusting constants related to millimeter-to-step conversions, step offsets, and Z-axis corrections until the machine accurately reached the desired coordinates.
\end{enumerate}

\subsection{Components Implementation}
The hardware system comprises the following primary hardware components:
\begin{itemize}
    \item \textbf{Structural Components:} All structural parts were 3D-printed using PLA filament with a gyroid infill pattern for optimal strength and material efficiency. The design includes the main frame, lead screw supports, pen holder, and motor mounts.
    \item \textbf{Microcontroller:} A Raspberry Pi Pico serves as the central control unit, selected for its cost-effectiveness and sufficient processing capability for real-time motor control.
    \item \textbf{Stepper Motors and Drivers:} Six 28BYJ-48 stepper motors, each paired with a ULN2003 driver board, control movement along the X, Y, and Z axes \cite{28BYJ48DataSheet,uln2003DriverDataSheet}. These motors offer adequate torque for the application's requirements while remaining inexpensive.
    \item \textbf{Lead Screws and Bearings:} Custom 3D-printed lead screws translate rotational motion from the motors into precise linear movement. 8\,mm $\times$ 22\,mm ball bearings support the lead screws, ensuring smooth operation and reducing friction.
    \item \textbf{Limit Switches:} CYT1073 limit switches are installed at the origin of each axis to support homing procedures and establish a consistent reference point.
    \item \textbf{Fasteners and Miscellaneous Components:} M3 and M4 screws of various lengths are used for assembly. The writing surface is constructed from foam board, providing a lightweight and inexpensive base for the paper.
\end{itemize}

\textbf{Mechanical Operation:}
Movement along each axis is achieved through the rotation of lead screws driven by the stepper motors. The pen holder travels through the X and Y axes to position the pen over the desired location, while the Z-axis controls the pen's contact with the paper. The system calculates the required number of motor steps based on the desired movement distance and the lead screw's pitch. Bearings secure the lead screws and reduce mechanical play, enhancing accuracy.

\begin{figure}[!htb]
    \centering
    \includegraphics[width=0.32\textwidth]{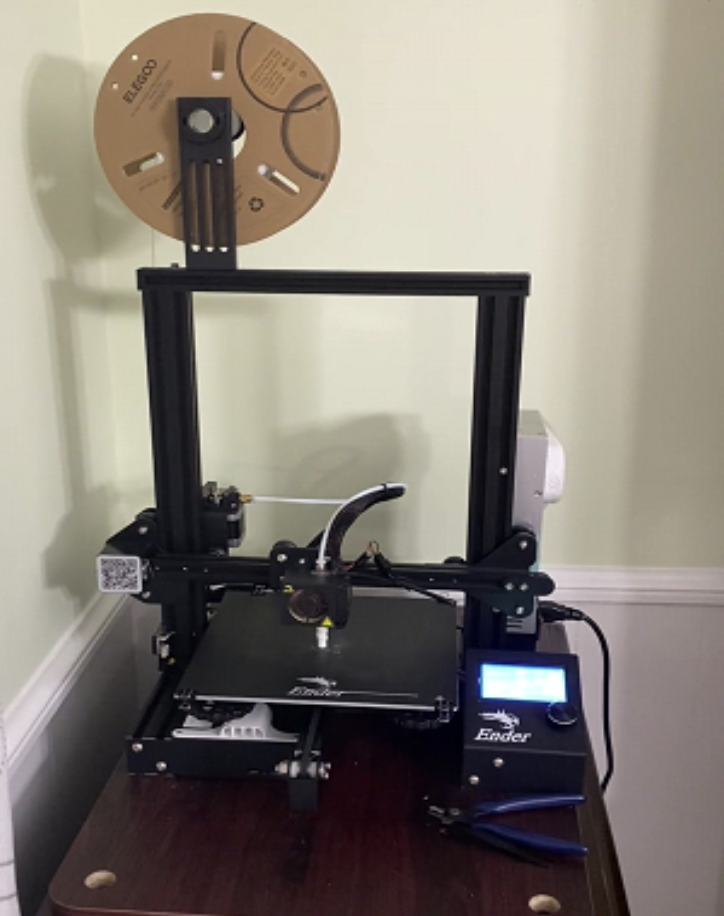}
    \caption{The Creality Ender 3 3D Printer.}
\end{figure}

\begin{figure}[!htb]
    \centering
    \includegraphics[width=0.48\textwidth]{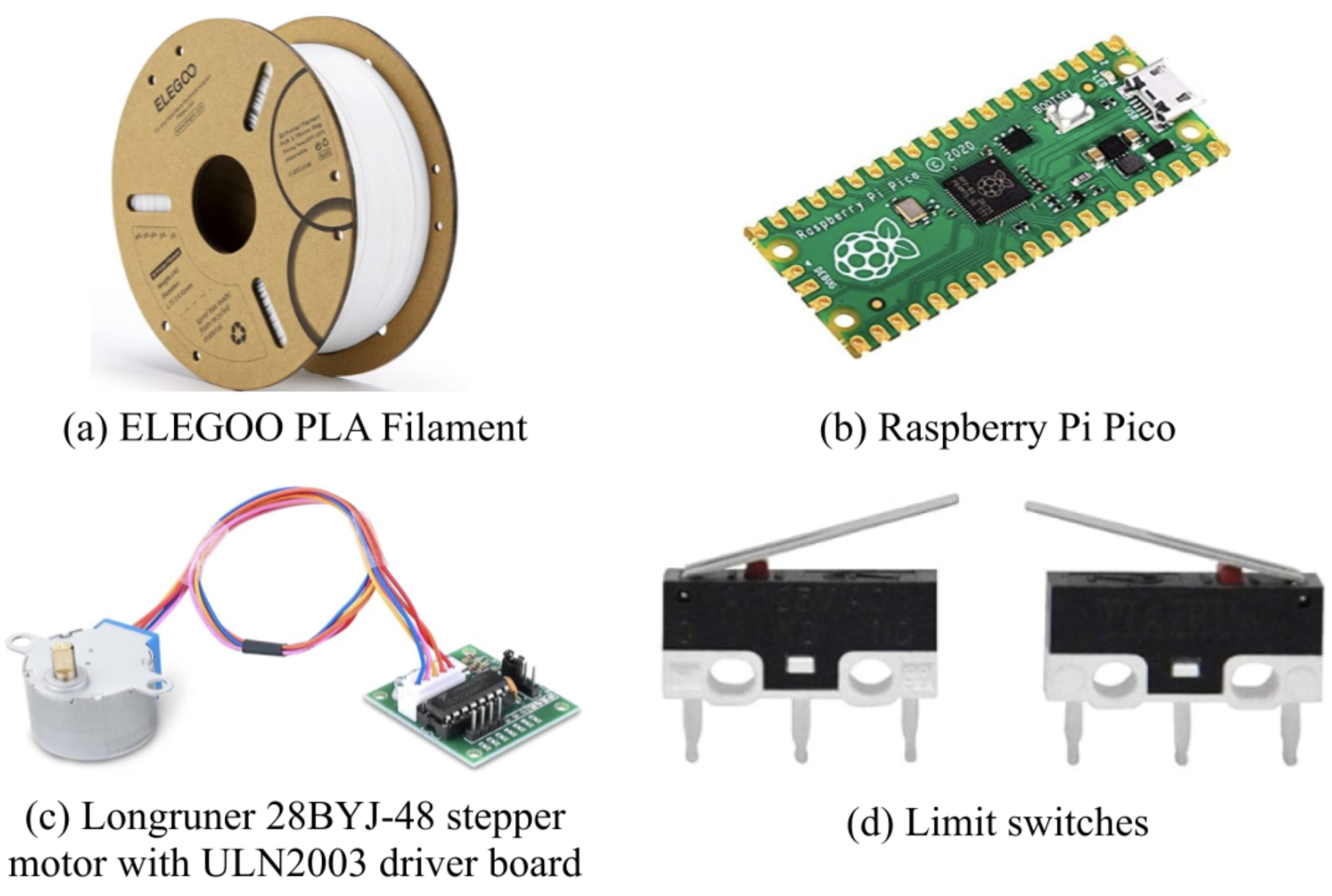}
    \caption{Main components used in the handwriting machine.}
\end{figure}

\begin{figure}[!htb]
    \centering
    \includegraphics[width=0.41\textwidth]{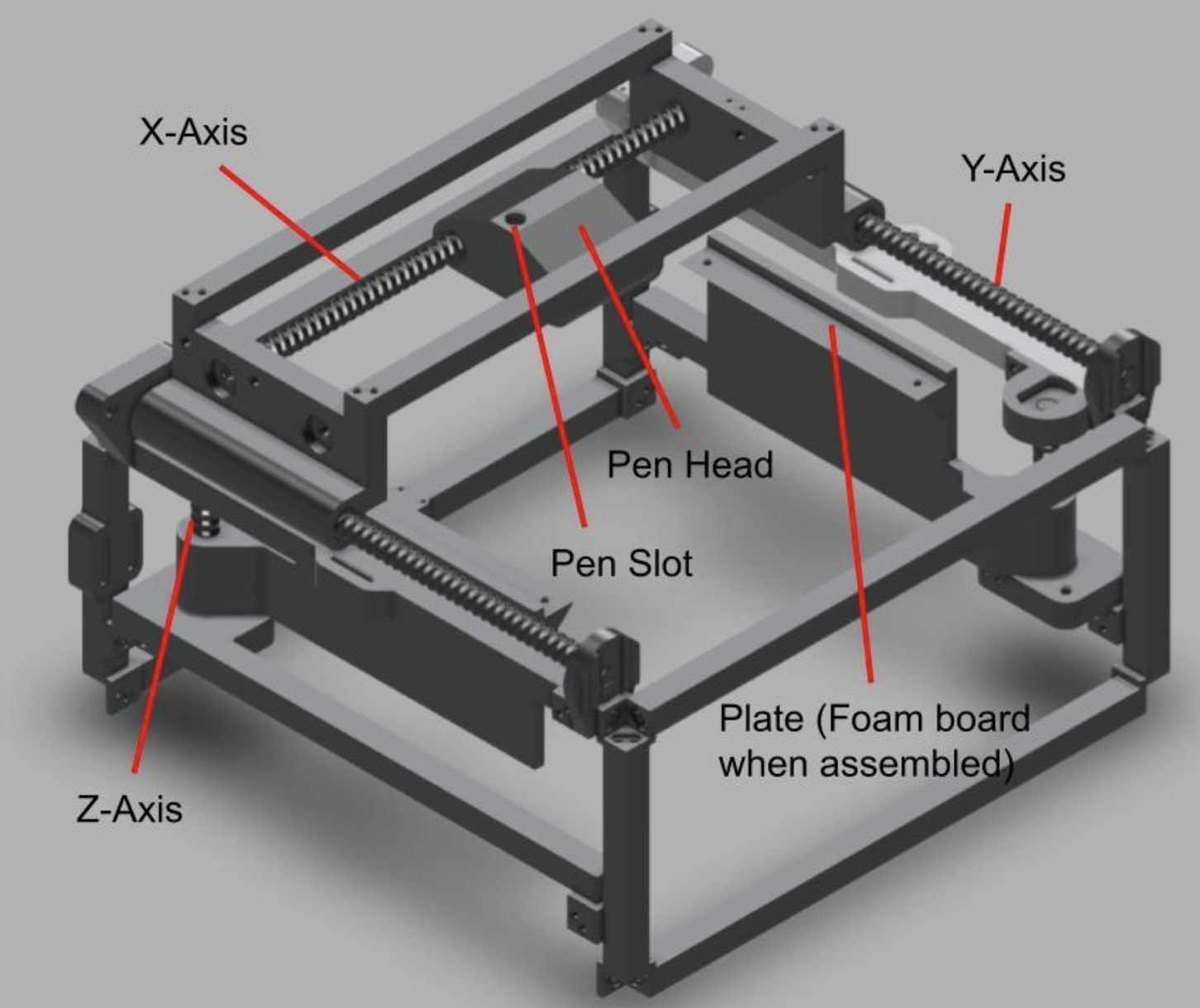}
    \caption{Labeled model of the frame, showing the arrangement of components and design.}
\end{figure}

\subsection{Control System Implementation}

To effectively control pen movements and execute motor commands rapidly, the system employs a dual-component control architecture comprising a microcontroller and a front-end application. The microcontroller, a Raspberry Pi Pico running an Arduino sketch, handles low-level motor control tasks. The front-end application, developed using JavaScript and TensorFlow.js \cite{malaviya2020tensorFlowjs}, manages high-level operations such as handwriting generation and coordinate transformation. These two components communicate over a direct USB serial connection.

\subsubsection{Front-End Application}
The front-end application serves as the user interface for handwriting generation. Its main functions include:
\begin{enumerate}
    \item \textbf{User Interaction and Initiation:} The user initiates the system by sending a "home" command, prompting the microcontroller to perform a homing procedure and position the machine at its origin. The user inputs the desired text into a textbox within the application.
    \item \textbf{Text Processing:} The input text is segmented into lines by determining the number of characters per line and splitting at spaces to avoid word fragmentation. The segmented text is converted into a series of coordinate points representing handwritten strokes using TensorFlow.js.
    \item \textbf{Coordinate Transformation:} The generated points, initially in millimeter units, are scaled to match the machine's coordinate system. This scaling factor is calculated based on the motor step size and lead screw pitch (e.g., scaling factor = steps per revolution divided by lead screw pitch). A constant offset is added to the X and Y coordinates to align them with the machine's reference frame.
    \item \textbf{Z-Axis Compensation:} An additional Z-axis offset is applied based on the pen's X and Y positions to compensate for mechanical flex and ensure consistent pen pressure across the writing area. This compensation uses a rectified linear unit (ReLU) function to adjust the pen's distance from the paper when it is farther from the Z-axis lead screws \cite{agarap2018deep}.
    \item \textbf{Pen-Up and Pen-Down Movements:} Coordinates are added to control the pen's lifting and lowering at the start and end of each stroke, ensuring proper stroke formation.
    \item \textbf{Data Management:} The transformed coordinates are organized into a queue for efficient management. The microcontroller communicates its buffer capacity to the front-end, which sends coordinates accordingly to prevent overloading the microcontroller.
\end{enumerate}

\subsubsection{Microcontroller Firmware}
The firmware on the Raspberry Pi Pico, written in C++ using the Arduino framework and a modified AccelStepper library \cite{arduinoIDE,accelStepperLib}, is responsible for executing precise motor control commands. Key aspects of the firmware include:
\begin{enumerate}
    \item \textbf{Initialization:} Configuring all input/output pins for communication with motors, drivers, and limit switches. Initializing AccelStepper objects for each motor.
    \item \textbf{Main Control Loop:} Continuously checks for incoming serial messages and determines if motor steps need to be executed. Processes received messages using a finite-state machine to execute the appropriate actions.
    \item \textbf{Homing Procedure:} Each axis moves toward its corresponding limit switch until triggered. After activation, the axis moves a predefined number of steps away from the switch to ensure accurate positioning. The current position is set to zero, establishing the origin (0, 0, 0).
    \item \textbf{Movement Execution:} Upon receiving target coordinates, the firmware calculates the required steps for each motor based on the difference between current and target positions. To achieve linear motion, the motor that needs to travel the greatest distance is set to its maximum speed. The speeds of the other motors are adjusted proportionally based on the ratio of their required distances to the longest distance.
    \item \textbf{Motor Control and Step Execution:} The AccelStepper library manages motor stepping with acceleration and deceleration profiles for smooth motion. Steps are executed until the target coordinates are reached.
    \item \textbf{Queue Management:} After completing a movement, the firmware updates its internal queue and sends feedback to the front-end about its capacity to receive more data. This ensures continuous operation without buffer overflows.
\end{enumerate}

\subsubsection{System Coordination}
Communication between the front-end application and the microcontroller uses a simple, custom protocol over the USB serial connection. The protocol includes:
\begin{itemize}
    \item \textbf{Command Messages:} The home command signals the microcontroller to perform the homing procedure. The move command contains target coordinate data for movement execution. Status updates are messages indicating movement completion and buffer capacity.
    \item \textbf{Data Formatting:} Messages are structured in a consistent format for reliable parsing. Example of a move command: \texttt{MOVE X:1200 Y:800 Z:200}
    \item \textbf{Feedback:} The microcontroller sends feedback on system status and readiness for new data upon receiving commands.
\end{itemize}

The system coordination benefits from this architecture in several ways. First, it ensures efficient resource usage by allowing the front-end to handle computationally intensive tasks, while the microcontroller focuses on time-sensitive motor control. Second, it provides flexibility as updates to the front-end software do not require changes to the firmware, and users can run the application on various computers with minimal requirements. Lastly, it enhances error handling by incorporating firmware checks for invalid commands and out-of-bounds coordinates, and by using limit switches and firmware safeguards to prevent mechanical overrun and damage.

\subsection{AI-Based Handwriting Generation}

A machine-learning handwriting generation model was developed using TensorFlow.js to process user-provided text and generate stroke trajectories, simulating natural handwriting. The main components of this AI system are:

\begin{enumerate}
    \item \textbf{Text Preprocessing:} The input text is segmented into individual letters. Each letter is extracted from the user-provided input to serve as the base for further processing.
    \item \textbf{Handwriting Model:} A pretrained recurrent neural network (RNN) model goes through post-training using samples of the user's handwriting to adapt to their style. This allows the system to generate coordinate sequences custom to the user's letter formation.
    \item \textbf{Stroke Generation:} Using the output of the fine-tuned RNN, the system generates precise pen movement coordinates for each letter \cite{Valle2019Stroke}.
    \item \textbf{Integration with Mechanical Control:} The generated coordinate data is passed to the control system, where it is translated into motor commands. This ensures that the machine physically reproduces the AI-generated handwriting accurately.
\end{enumerate}

\section{Results and Discussions}

\subsection{System Performance}
To assess precision, the machine was programmed to write lines of text generated by the AI handwriting model. The output was compared to the original by overlaying the machine's writing onto a printed version of the AI-generated text under bright light \cite{Carrillo2020CNCAccuracy}. The maximum deviation between corresponding points was measured using a caliper.

An example of this comparison is shown in Figure~6. The green line represents the measured deviation between a point on the machine's output and the target point. After testing 10 full lines of text, the maximum observed deviation was within $\pm$0.3 millimeters, indicating sufficient accuracy. The maximum speed of the machine was also measured to be approximately 200~mm/min using calipers and a stopwatch.

\begin{figure}[!htb]
    \centering
    \includegraphics[width=0.3\textwidth]{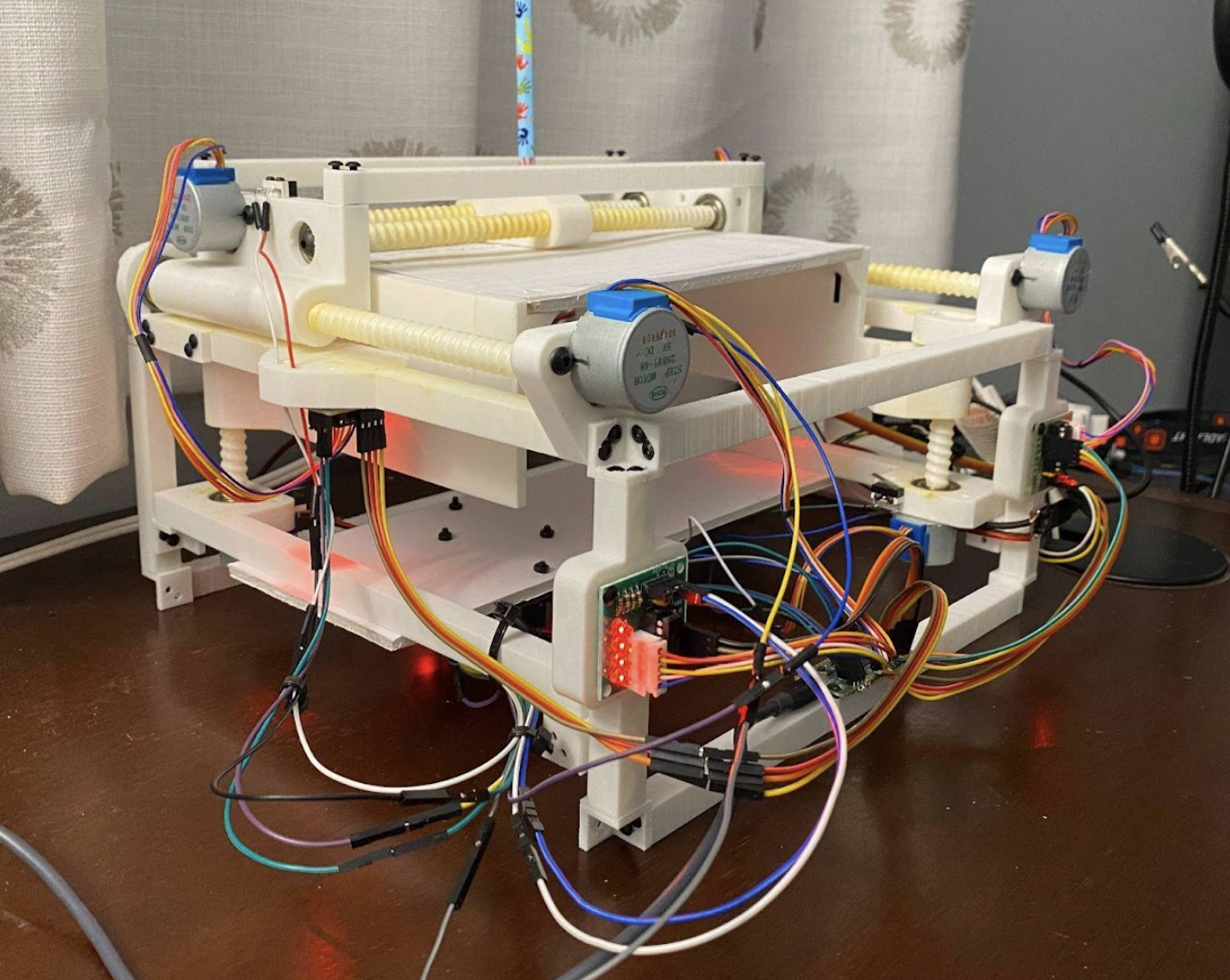}
    \caption{The fully assembled handwriting machine equipped with a pencil.}
\end{figure}

\begin{figure}[!htb]
    \centering
    \includegraphics[width=0.3\textwidth]{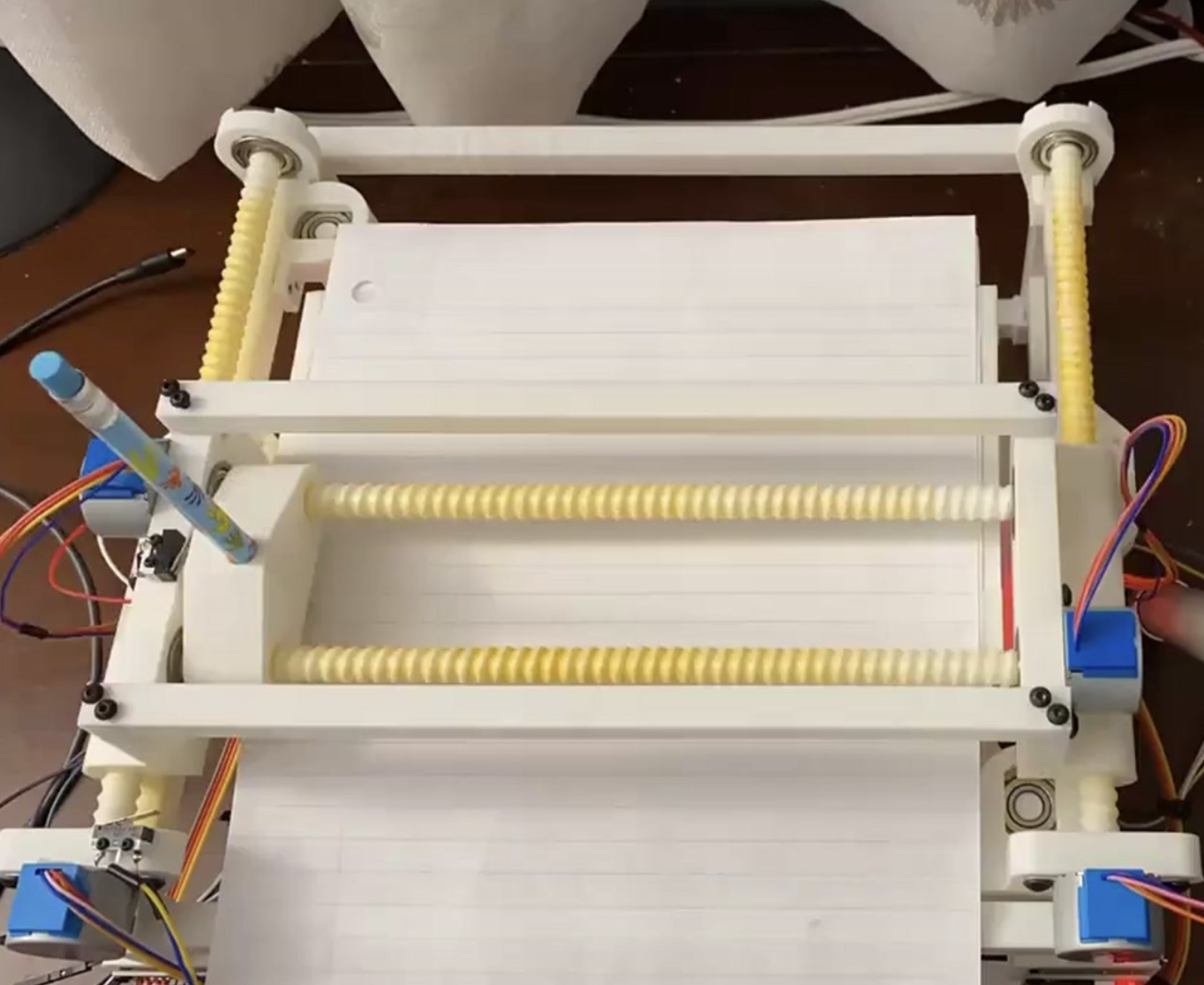}
    \caption{The machine writing on a piece of paper.}
\end{figure}

\begin{figure}[!htb]
    \centering
    \includegraphics[width=0.36\textwidth]{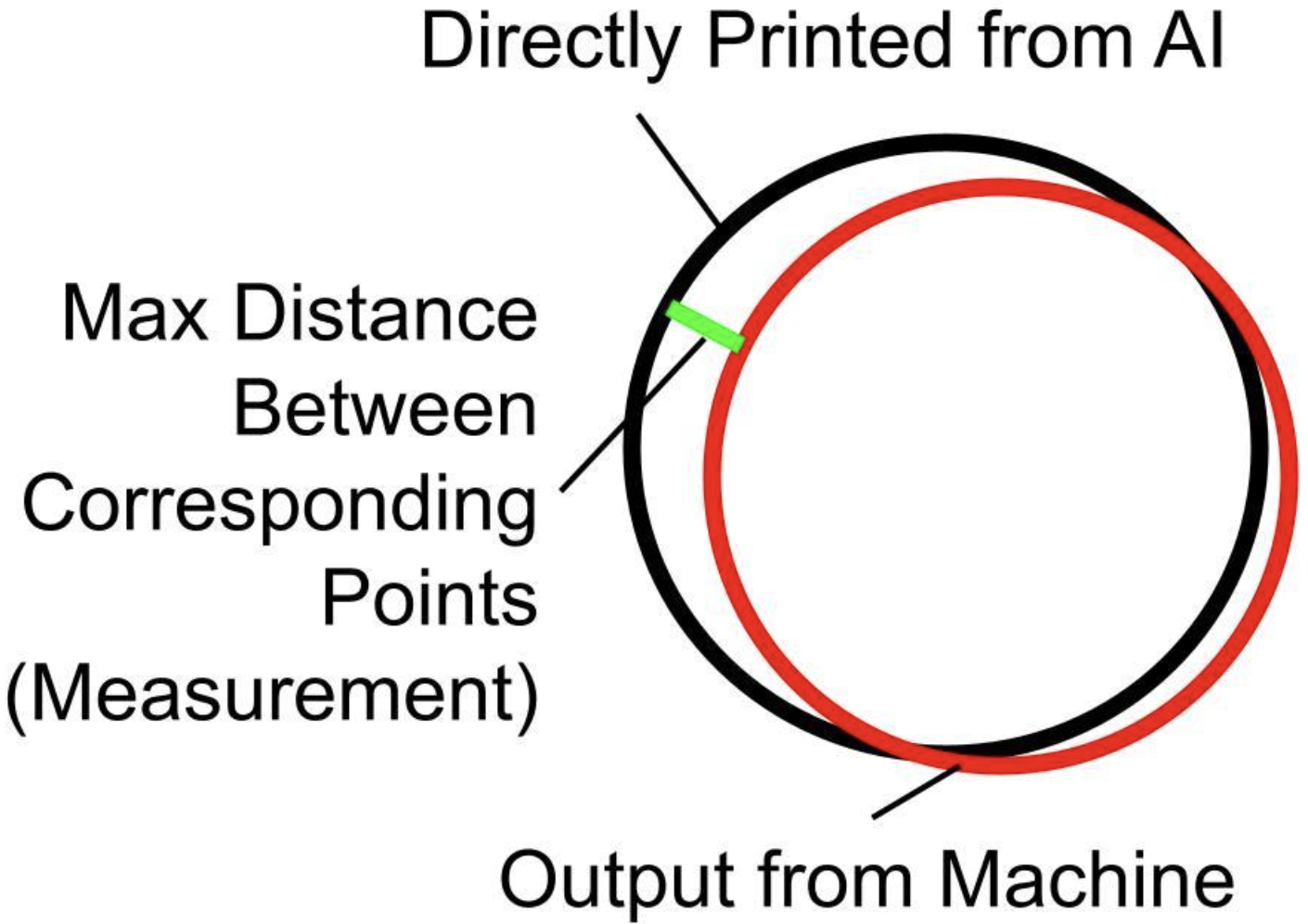}
    \caption{Demonstration of the accuracy testing method with overlaid images and highlighted deviation.}
\end{figure}

\begin{figure}[!htb]
    \centering
    \includegraphics[width=0.3\textwidth]{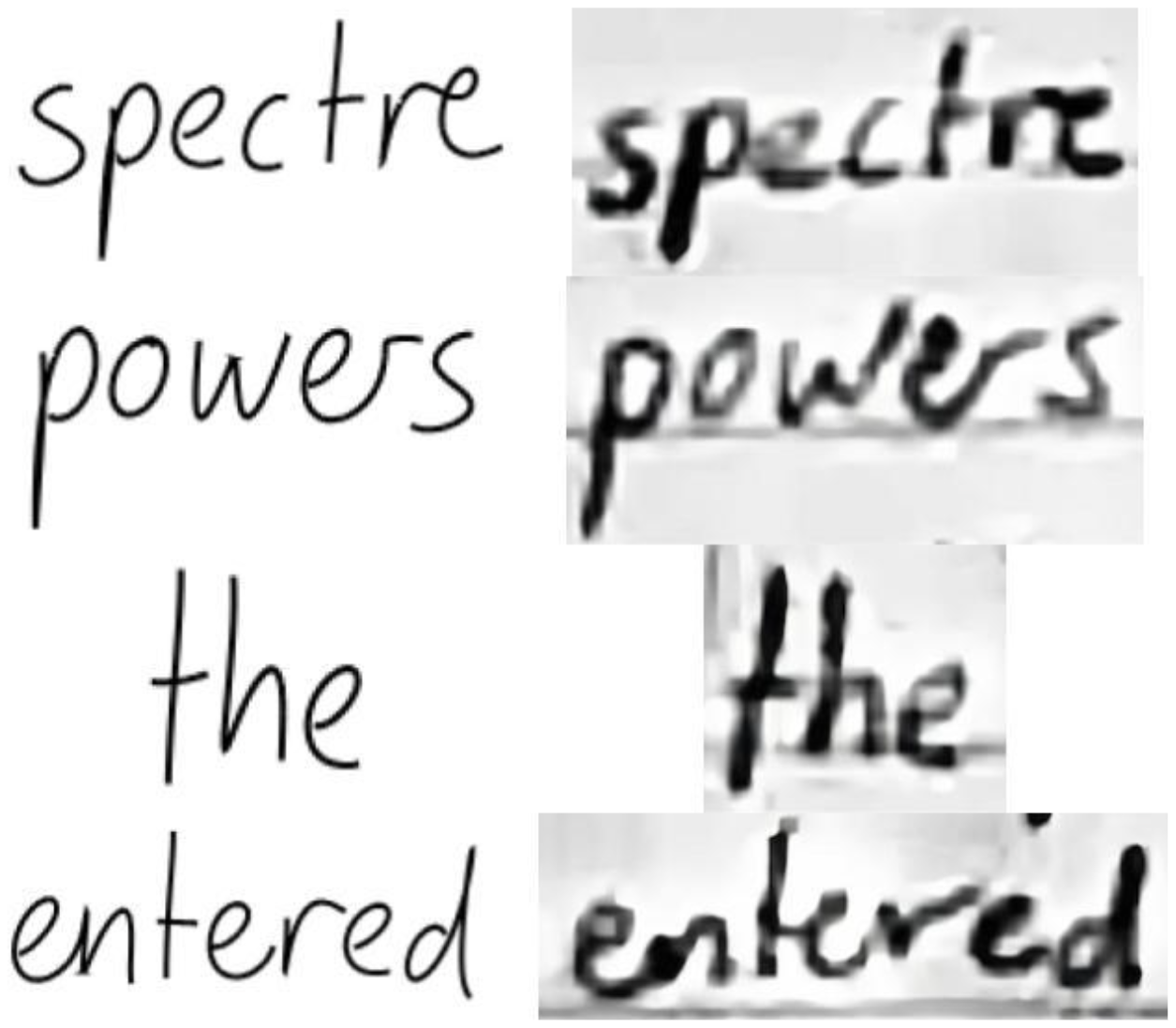}
    \caption{Side-by-side comparison of the AI-generated handwriting (left) and the machine's reproduced handwriting (right).}
\end{figure}

\subsection{Cost Analysis}
A complete cost analysis was conducted to compare this machine with existing commercial handwriting machines. The total cost of the primary components is outlined in Table~\ref{tab:cost-breakdown}.

\begin{table}[!htb]
    \centering
    \caption{Cost Breakdown of the Handwriting Machine Components}
    \begin{tabular}{|l|c|c|}
        \hline
        \textbf{Component} & \textbf{Count} & \textbf{Total Cost (USD)} \\
        \hline
        Raspberry Pi Pico & 1 & 4.00 \\
        Motors/Drivers & 6 & 14.00 \\
        Bearings & 12 & 5.40 \\
        Foam Board & 1 & 3.00 \\
        Screw Set & 1 & 13.00 \\
        Limit Switch & 3 & 0.69 \\
        Filament & 1 & 10.00 \\
        Jumper Wires & 1 & 6.00 \\
        \hline
        \textbf{Total} & & \textbf{56.09} \\
        \hline
    \end{tabular}
    \label{tab:cost-breakdown}
\end{table}

The handwriting machine's cost-effectiveness results from inexpensive material selections and design simplifications that greatly reduce expenses without compromising performance. One primary factor is the substitution of traditional materials with more affordable alternatives. Conventional handwriting machines often use aluminum chassis to enhance stability, which is necessary for devices like 3D printers and CNC routers that handle heavier loads. However, moving a single pen or pencil does not require such robust construction. Therefore, this machine replaces aluminum rails with 3D-printed plastic components with gyroid infill patterns, which provide sufficient strength while minimizing material usage \cite{Rodrigues2019Gyroid}.

Another significant cost-saving measure is the simplified movement system. Traditional machines employ timing-belt systems, which involve additional expenses for belts, pulleys, and tensioning mechanisms. This machine adopts 3D-printed lead screws for movement along each axis, a method commonly used for Z-axis motion in 3D printers due to its precision and strength. Lead screws can be entirely produced using a 3D printer, eliminating the need to purchase costly belts and pulleys, and simplifying assembly by removing the need for complex tensioning.

Furthermore, the mechanical advantage of the lead screw system enables the use of less powerful and more affordable stepper motors. Instead of the typical NEMA~17 motors found in many commercial machines, we opted for 28BYJ-48 stepper motors paired with ULN2003 driver boards. These motors meet the system's reduced load and precision requirements, yielding additional savings.

Overall, these cost-saving measures—using a simplified movement system and affordable motors/drivers—reduce the total machine cost by approximately \$100 compared to commercial alternatives. This reduction allows for better accessibility, making advanced robotic handwriting technology available to a broader audience, including individuals and institutions with limited budgets.

\subsection{Advantages and Limitations}
With its modified structure and movement system, this handwriting machine presents clear advantages while also facing specific limitations that warrant consideration for future development.

\textbf{Advantages:}
\begin{enumerate}
    \item \textbf{Accessibility:} By reducing the cost to about one-third of existing commercial machines, it becomes accessible not only to institutions but also to individuals, including those with disabilities who may benefit from assistive writing devices.
    \item \textbf{Simplified Assembly and Customization:} The integration of a lead screw system simplifies the assembly process by eliminating the need for belt tensioning and complex alignments. Additionally, the use of 3D-printed components allows users to easily customize and prototype parts.
    \item \textbf{Material Efficiency:} 3D-printed plastics reduce material waste and energy consumption compared to machining metal components. The ability to produce parts on demand also contributes to a more sustainable manufacturing approach \cite{Kumar2020Sustainable3D}.
\end{enumerate}

\textbf{Limitations:}
\begin{enumerate}
    \item \textbf{Reduced Writing Speed:} The use of more cost-effective stepper motors and low-pitch lead screws results in slower operation compared to other handwriting machines. This may reduce productivity when handling larger volumes of text.
    \item \textbf{Longer Assembly Time:} Since many structural components must be 3D-printed by the user and electronic parts purchased separately, the assembly process will likely be more time-consuming and complex for individuals without prior experience.
    \item \textbf{Limited Writing Volume:} The size constraints of common 3D printers limit the dimensions of the machine's components, resulting in a smaller writing area compared to commercial machines. For handwriting full pages of text, the machine may require multiple passes or repositioning of the paper.
\end{enumerate}

\subsection{Comparative Study}
To analyze the advancements of the handwriting machine, we compare it with two existing systems: an Arduino-based 3-axis plotter machine by Hasan et al. and a handwriting robot by Zamani et al.

Hasan et al. developed a low-cost plotter machine using an Arduino and CNC shield. Their machine operates on three axes (X, Y, and Z) with stepper motors controlled by an Arduino microcontroller. The movement ranges are 215~mm on the X-axis and 235~mm on the Y-axis. The system is modeled using SolidWorks and relies on standard CNC principles \cite{Hasan2021LowCostPlotter}. However, the paper does not detail specific strategies for cost reduction beyond utilizing open-source hardware and software.

Zamani et al. presented a handwriting robot that also uses an Arduino microcontroller to control stepper motors along the X and Y axes, with a servo motor managing the pen's up-and-down movement on the Z-axis. They employ Inkscape and G-Code for signal generation from drawings, enabling the replication of intricate graphics and text \cite{Zamani2022HandwritingRobot}. Their focus is on creating a system capable of handling both graphics and handwriting using standard CNC methodologies.

In contrast, the handwriting machine introduces several innovations to enhance cost-effectiveness and accessibility. It employs a Raspberry Pi Pico microcontroller, offering sufficient processing power for real-time motor control at a lower cost. The system uses six 28BYJ-48 stepper motors paired with ULN2003 driver boards for precise control along all three axes, optimizing both performance and affordability.

Moreover, the handwriting machine extensively uses 3D-printed components made from PLA filament with gyroid infill patterns. This choice significantly reduces material costs and allows for easy customization and prototyping. By replacing traditional metal components with lightweight plastic parts, this machine achieves a reduction in overall weight and manufacturing expenses.

Another innovation is the adoption of 3D-printed lead screws for movement along all axes, replacing conventional timing belts and pulleys. This simplification reduces mechanical complexity, eases assembly, and enhances precision by eliminating the need for belt tensioning.

In terms of cost, the handwriting machine stands out with a total estimated cost of approximately \$56 per unit. This is achieved through strategic material selection, simplified mechanical design, and the use of affordable components. While the exact costs of the systems by Hasan et al. and Zamani et al. are not specified, their reliance on standard CNC components and metal structures suggests higher expenses.

Performance-wise, the handwriting machine demonstrates handwriting precision within $\pm$0.3\,mm and effectively reproduces AI-generated handwriting. The other systems, based on their studies, focus on general plotting and handwriting capabilities but do not detail precision metrics.

In summary, the handwriting machine differentiates itself by offering a cost-effective, customizable, and precise robotic handwriting solution. By leveraging affordable materials and simplified mechanics, it addresses the limitations of existing systems and expands accessibility for applications requiring personalized handwriting production.

\section{Conclusion and Future Work}
This paper presented the design and implementation of a cost-effective robotic handwriting system utilizing a Raspberry Pi Pico microcontroller and 3D-printed components. By simplifying mechanical designs and using affordable materials, the system reduces the overall cost to approximately \$56---significantly lower than the \$150 cost of existing commercial machines. Despite the reduced cost, the handwriting machine maintains a high accuracy in handwriting reproduction, achieving a maximum deviation of $\pm$0.3\ millimeters.

The successful development of this machine demonstrates that advanced robotic handwriting technology can be made accessible to a wider audience without sacrificing performance. The use of 3D-printed components and a lead screw mechanism not only reduces costs but also simplifies assembly and allows for greater customization, making it suitable for various applications, including education and assistive technologies.

\subsection{Future Work}
Future enhancements of the handwriting machine could focus on improving writing speed by experimenting with higher-pitch lead screws or upgrading motors. Structural improvements to increase rigidity might further enhance accuracy and eliminate the need for software compensation. Simplifying the assembly process through the development of a custom PCB or offering pre-assembled kits could make the system more user-friendly. Additionally, expanding the writing area by designing modular components or integrating a paper-feeding mechanism would increase functionality. Adapting the system to accommodate various writing instruments could broaden its applications to include artistic expressions such as calligraphy.

\bibliographystyle{IEEEtran}
\bibliography{ms}

\end{document}